\documentclass[10pt,twocolumn,letterpaper]{article}

\usepackage{iccv}
\usepackage{times}
\usepackage{epsfig}
\usepackage{graphicx}
\usepackage{amsmath}
\usepackage{amssymb}
\usepackage{graphicx}

\usepackage{booktabs}
\usepackage{xcolor}
\usepackage{multirow}

\usepackage{caption}
\usepackage{subcaption}
\usepackage[pagebackref=true,breaklinks=true,letterpaper=true,colorlinks,bookmarks=false]{hyperref}

\iccvfinalcopy 

\usepackage{enumitem}
\usepackage{float}

\def\iccvPaperID{} 

\begin{document}

\title{Improved Diffusion-based Image Colorization via Piggybacked Models}

\author{
\begin{tabular}{ccccc}
Hanyuan Liu\textsuperscript{\rm 1} &
Jinbo Xing\textsuperscript{\rm 1} &
Minshan Xie\textsuperscript{\rm 1} &
Chengze Li\textsuperscript{\rm 2} &
Tien-Tsin Wong\textsuperscript{\rm 1}\thanks{Corresponding author.}
 \end{tabular}\\
\begin{tabular}{cc}
\textsuperscript{\rm 1} The Chinese University of Hong Kong & 
\textsuperscript{\rm 2} Caritas Institute of Higher Education\\
 \end{tabular}\\
\begin{tabular}{cc}
{\tt\small \{liuhy, jbxing, msxie, ttwong\}@cse.cuhk.edu.hk} &
{\tt\small czli@cihe.edu.hk}
 \end{tabular}
}

\maketitle
\ificcvfinal\thispagestyle{empty}\fi

\begin{abstract}
Image colorization has been attracting the research interests of the community for decades. However, existing methods still struggle to provide satisfactory colorized results given grayscale images due to a lack of human-like global understanding of colors. Recently, large-scale Text-to-Image (T2I) models have been exploited to transfer the semantic information from the text prompts to the image domain, where text provides a global control for semantic objects in the image. In this work, we introduce a colorization model piggybacking on the existing powerful T2I diffusion model. Our key idea is to exploit the color prior knowledge in the pre-trained T2I diffusion model for realistic and diverse colorization. A diffusion guider is designed to incorporate the pre-trained weights of the latent diffusion model to output a latent color prior that conforms to the visual semantics of the grayscale input. A lightness-aware VQVAE will then generate the colorized result with pixel-perfect alignment to the given grayscale image. Our model can also achieve conditional colorization with additional inputs (e.g. user hints and texts). Extensive experiments show that our method achieves state-of-the-art performance in terms of perceptual quality.

\end{abstract}
\newcommand*{\approxident}{%
  \mathrel{\vcenter{\offinterlineskip
  \hbox{$\sim$}\vskip-.35ex\hbox{$\sim$}\vskip-.35ex\hbox{$\sim$}}}}
\newcommand{\RR}{\mathbb{R}}
\newcommand{\expec}{\mathbb{E}}
\newcommand{\prior}{\mathcal{N}(0,1)}

\newcommand{\PriorModel}{latent diffusion guider model}
\newcommand{\ColorizationDecoder}{lightness-aware VQVAE model}

\newcommand{\xpixel}{x_\text{pixel}}
\newcommand{\xpixelrec}{\tilde{x}_\text{pixel}}
\newcommand{\hpixel}{H}
\newcommand{\wpixel}{W}
\newcommand{\cpixel}{3}
\newcommand{\xrec}{\tilde{x}}

\newcommand{\latent}{z_0}
\newcommand{\hlatent}{h}
\newcommand{\wlatent}{w}
\newcommand{\clatent}{c}

\newcommand{\zt}[1]{z_{#1}}
\newcommand{\enc}{E}
\newcommand{\ppixel}{p_\text{pixel}}
\newcommand{\pzt}[1]{p_{\zt{#1}}}
\newcommand{\qzt}[1]{q_{\zt{#1}}}
\newcommand{\q}{q}
\newcommand{\R}{\mathbb{R}}

\newcommand{\LPIPS}{\text{LPIPS}}
\newcommand{\KL}{\mathbb{KL}}
\newcommand{\expect}{\mathbb{E}}
\newcommand{\pmodel}[1]{p^{#1}_{\theta}}
\newcommand{\pchain}{p_{\theta}}
\newcommand{\qchain}{q}
\newcommand{\qmodel}[1]{q_{#1}}

\newcommand{\qenc}{q_{\phi}}
\newcommand{\pdec}{p_{\phi}}
\newcommand{\dec}{G_{\phi}}

\newcommand{\disc}{D_{\psi}}

\newcommand{\lrec}{L_{rec}}
\newcommand{\ladv}{L_{adv}}
\newcommand{\lreg}{L_{reg}}
\newcommand{\lcomp}{L_{cm}}
\newcommand{\lsimple}{L_{DM}}
\newcommand{\lsimpleldm}{L_{LDM}}
\newcommand{\lsimplelcm}{L_{LDM}}
\newcommand{\lcolor}{L}

\newcommand{\model}{\epsilon_\theta}
\newcommand{\conditioner}{\tau_\theta}

\newcommand{\encoder}{\mathcal{E}}
\newcommand{\GrayEncoder}{{\mathcal{G}}}
\newcommand{\decoder}{\mathcal{D}}

\newcommand{\cond}{y}

\newcommand{\guider}{\mathcal{F}}

\def\headline#1{\hbox to \hsize{\hrulefill\quad\lower.3em\hbox{#1}\quad\hrulefill}}

\section{Introduction}

Image colorization aims to assign colors to grayscale images, enabling us to bring the legacy photographs to life. However, this task is challenging, as we need to strike a balance among the global color richness, the visual harmony, and the conformity to object-level semantics, while minimizing the visual artifacts such as color leakage or incomplete colorization. 
These challenges have been studied for decades in computer vision and graphics, including traditional methods \cite{welsh2002transferring} and learning-based solutions \cite{zhang2017real,kumar2021colorization}. 

Recent diffusion-based text-to-image (T2I) models \cite{dalle2,latent-diffusion} have emerged as a powerful technique for generating highly realistic and visually plausible images. They leverage the ability of {zero-shot learning \cite{openai-clip}} to transfer the semantic information from the text prompts to the image domain.
Recent works~\cite{sketch-diffusion, control-net, t2iadapter} demonstrate that diffusion-based T2I models have the potential to better understand and handle semantics. It would be nice if we could exploit such a semantic understanding potential to address the challenges of image colorization. Moreover, the T2I models provide a more natural way, {\em text}, for nonprofessional users to specify the desired color composition and object colors. 
Such text provides a strong supplement when grayscale image features are insufficient for complete color inference. In other words, we can formulate image colorization as a T2I problem. Instead of solely feeding text to the T2I models for image generation, we can feed the grayscale image and the text description of color composition to the model for generating the desired colorized output.
However, existing pretrained T2I models lack the ability to condition on images to regulate the output to be aligned with the condition, making it impossible to achieve our colorization goal with simple finetuning. Additionally, the training cost of a customized T2I model from scratch is extremely high \cite{latent-diffusion}. 

In this work, we propose a new diffusion-based model that works across the latent and pixel domain for vivid and perceptually consistent grayscale image colorization. Different from the synthesis-only T2I models \cite{dalle2,latent-diffusion, saharia2022photorealistic}, our colorization pipeline allows the grayscale input and optional user color hints as immediate conditions to guide the denoising diffusion process. While our pipeline exhibits a different nature and philosophy, we still aim to \emph{piggyback} on the existing Stable Diffusion model \cite{latent-diffusion}, to incorporate its demonstrated strong capabilities in image semantic understanding and synthesis into our pipeline. 
Specifically, the piggybacking involves two main components: (1) adapting the pre-trained Stable Diffusion weights for the newly formulated grayscale colorization problem, and (2) adapting to transit from the latent-level feature colorization into exact pixel-level colorization.

Specifically, we first propose a diffusion guider model which lies upon a pretrained diffusion model to adapt and route its outstanding feature learning ability. The diffusion guider parses the features from the grayscale input to generate a latent color prior that conforms to the input lightness information. The latent prior is afterwards decoded via a lightness-aware VQVAE decoder that shortcuts the information from the grayscale input to depict pixel-perfect alignment of structure and texture details in the colorization, which is unable to be achieved in the original Stable Diffusion model. In addition, our colorization pipeline is feasible to generate conditioned and diverse colorized images by providing additional user annotations. 

Our proposed model, which piggybacks on the strengths of the Stable Diffusion model while incorporating additional conditional inputs, offers a more efficient and environmentally friendly solution to the challenges of grayscale image colorization. By leveraging the power of a large-scale pretrained model, our approach can outperform a colorization model trained solely on our own dataset, while using significantly less data and computational resources.

We summarize our contribution as follows:
\begin{itemize}[nolistsep]
    \item We propose a novel pipeline that piggybacks on the pre-trained latent diffusion models for realistic and diverse image colorization with a much reduced requirement in training data and power. 
    \item We design a diffusion guider model that incorporates the pretrained weights of latent diffusion model to output a latent color prior that conforms to the visual semantics of the grayscale input.
    \item We propose a novel lightness-aware VQVAE model that shortcuts the grayscale information to the decoder for pixel-perfect aligned colorization against the input. 
\end{itemize}

\section{Related Work}

\textbf{User-assisted Colorization.}
Early colorization methods utilize user-annotated colors and propagate them to nearby pixels based on their similarity~\cite{qu2006manga,yatziv2006fast,cho2018text2colors}. Local hints, such as user scribbles~\cite{levin2004colorization}, and global hints, such as color palette ~\cite{chang2015palette} or text\cite{l-code}, can be useful annotations to guide colorization. The color hints are then propagated to neighbor pixels based on pixel affinity by solving constrained optimization problems~\cite{levin2004colorization,qu2006manga,endo2016deepprop}. With the advent of deep learning, ~\cite{zhang2017real} proposed a real-time user interactive colorization model which can generate visually plausible results with sparse color hints. 
However, these methods may fail to provide robust and automatic colorization when dense user annotations are not available. 

\textbf{Reference-based Colorization}
methods colorize the target grayscale image by transferring color information from a reference image using global color ~\cite{welsh2002transferring} or some local similarity ~\cite{bugeau2013variational}. Recently, some researchers~\cite{he2019progressive, xu2020stylization} have made use of deep features extracted by a pretrained neural network for reliable spatial correspondence and generate plausible results. However, these methods usually require a content-related image as reference, which poses challenges to users. To alleviate this obstacle, \cite{he2018deep} propose to utilize natural color distributions learned from large-scale image data when no appropriate matching region is available in the reference. 

\textbf{Automatic Colorization}
 is generally a kind of learning-based method, which uses neural networks to learn the mapping of grayscale images to color images~\cite{cheng2015deep, deshpande2015learning, isola2017image}. These methods can be trained on large-scale datasets and do not require user input or reference images. Some follow-up works promote the colorized results by integrating additional semantic guidance, like class labels~\cite{larsson2016learning, iizuka2016let}, semantic segmentation maps ~\cite{larsson2017pixcolor,zhao2020pixelated} and instance-level bounding box ~\cite{su2020instance}. 
To improve the diversity of the colors generated, \cite{zhang2016colorful} proposed a method that uses a convolutional neural network (CNN) to predict a probability distribution over the possible colors for each pixel. Generative Adversarial Network (GAN)~\cite{goodfellow2014generative} is adopted for realistic and vivid colorization~\cite{vitoria2020chromagan}. The transformer model is also used for various high-fidelity image colorizations~\cite{kumar2021colorization,ji2022colorformer,color-token}. 
Recently, some attempts have attempted to directly predict sparse guidance from the grayscale image for controllable and diverse colorization, such as color palette\cite{wang2022palgan}, local color hints\cite{disco-color}, etc. \cite{unicolor} propose a colorization framework for incorporating various conditions, such as stroke, exemplar, text, or a combination of them, to guide the colorization.
However, these methods still suffer from visual artifacts like unnatural and incoherent colors for most man-made scenes that are generally with color uncertainty. 

\textbf{Colorization with Generative Priors.} 
For realistic and vivid colorization, generative priors of pretrained GANs~\cite{karras2019style,karras2020analyzing} have recently been exploited by GAN inversion~\cite{zhu2016generative,zhu2020domain,abdal2019image2stylegan}. These methods first invert the grayscale to a latent code of the pretrained GAN, and then reconstruct the color image with iterative optimization. 
Deep generative prior (DGP)~\cite{pan2020exploiting} jointly optimizes the latent code and the pretrained generator to synthesize a color image via GAN inversion. However, these methods struggle to synthesize faithful results, since low-dimensional latent codes without spatial information are insufficient to guide colorization. To achieve high texture faithfulness, \cite{wu2021towards} propose to utilize GAN inversion techniques to first obtain a content-related image and then warp the synthesized color features into the grayscale image. \cite{kim2022bigcolor} enlarges the representation space by using an encoder-generator architecture that uses spatial features instead of a spatially-flattened latent code.
But, the colorized results usually rely on the quality of images generated by GAN inversion.

\section{Method}
\begin{figure*}
\centering
\includegraphics[width=1.0\linewidth]{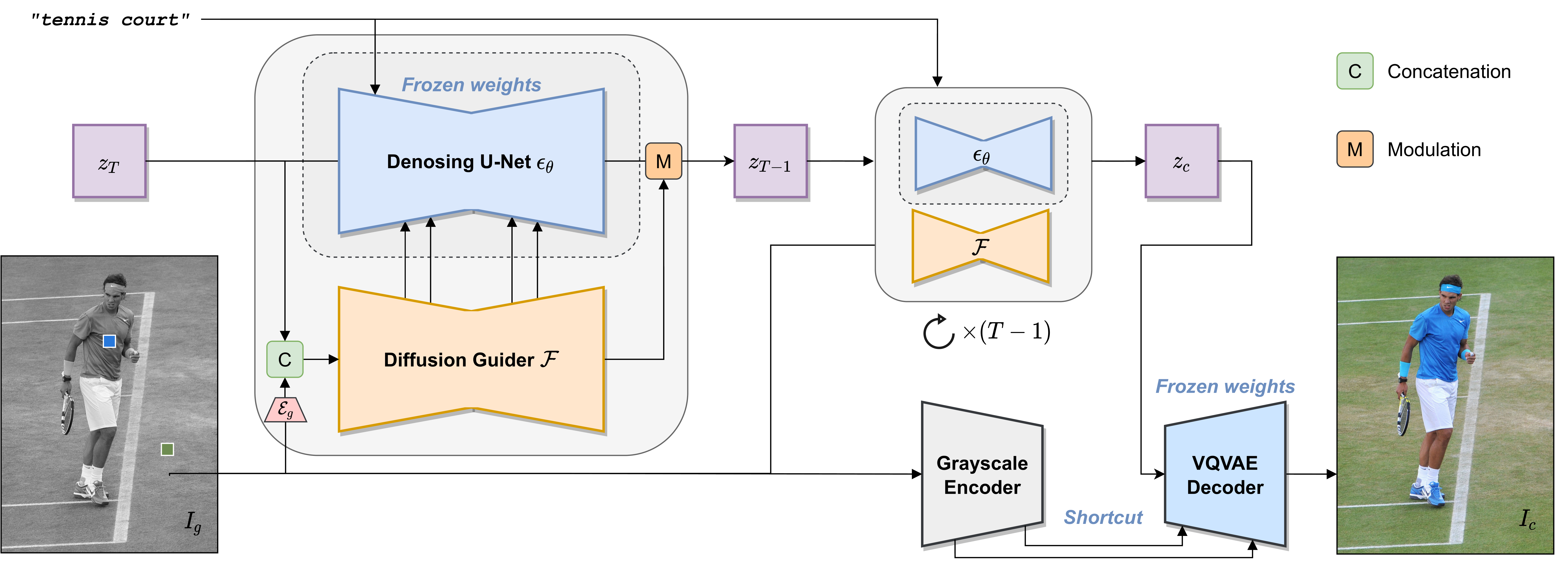}
   \caption{Overview. Our framework consists of two components: a \PriorModel and a \ColorizationDecoder. When the grayscale image $I_g$ and the optional textual descriptions $t$ or hint points $\{h\}$, then the \PriorModel~guides the pretrained Stable Diffusion model to generate a ``colorized" latent $z_c$ through the denoising diffusion process. The \ColorizationDecoder~then uses $z_c$ as the latent color prior and incorporates the grayscale information of $I_g$ to produce a pixel-aligned colorization $I_c$.}
\label{fig:overview}
\end{figure*}
\subsection{Overview}

In this work, we propose a pipeline for grayscale images to produce high-quality and diverse colorization, conditioned on textual descriptions and color hint points. Specifically, the pipeline is built upon the pretrained Stable Diffusion model to incorporate its capability in visual semantic understanding and synthesis. We illustrate the overall pipeline in Figure~\ref{fig:overview}. 
The pipeline consists of two main building blocks. The first is a \PriorModel~$\mathcal{F}$ that leverages the power of the Stable Diffusion model to render colors in the latent space $z_c$, according to the visual semantics of the grayscale input $I_g$. Additionally, the \PriorModel~allows user inputs such as text prompts or color hints for better control of the colorization. The second building block is the novel \ColorizationDecoder~$\decoder$ that fuses the compressed VQ color prior $z_c$ and the grayscale input $I_g$ to reconstruct the final colorization $I_c$ that perfectly aligns with the structure and texture of the grayscale input $I_g$.

\subsection{Latent Color Prior Diffusion Model}
\subsubsection{Preliminary: latent diffusion models}
\textbf{Diffusion Models} \cite{diffusion-model} are probabilistic models designed to learn a data distribution $p(x)$ by gradually denoising a normally distributed variable. This denoising process or the generation process corresponds to learning the reverse process of a fixed Markov Chain of length $T$ \cite{ddpm, openai-diffusion, sr3-diffusion}. For image synthesis, the most successful models rely on a reweighted variant of the variational lower bound on $p(x)$, which mirrors denoising score-matching \cite{score-sde}. These models can be interpreted as an equally weighted sequence of denoising autoencoders $\model(x_{t},t);\, t=1\dots T$, which are trained to predict a denoised variant of their input $x_t$, where $x_t$ is a noisy version of the input $x$. Instead of directly learning the target data distribution in the high-dimensional pixel space, \textbf{Latent Diffusion Models} \cite{latent-diffusion} leverage perceptual compression with an autoencoder $\encoder$ and $\decoder$ for efficient low-dim representations features. The model learns $p(z|y)$, in which $\encoder(x)=z, \encoder(\decoder(x))\approx x$ for reconstruction:
\begin{equation}
\lsimplelcm = \expec_{\encoder(x), y, \epsilon \sim \mathcal{N}(0, 1), t }\Big[ \Vert \epsilon - \model(z_{t},t, y) \Vert_{2}^{2}\Big] \, ,
\label{eq:cond_loss}
\end{equation}
with $t$ uniformly sampled from $\{1, \dots, T\}$. The neural backbone $\model(\circ, t, \conditioner(y))$ is generally realized as a denoising U-Net~\cite{unet} with cross-attention conditioning mechanisms \cite{attention} to accept additional conditions. 

\subsubsection{Text-to-image latent diffusion model as color prior}

The recent state-of-the-art text-to-image diffusion models \cite{latent-diffusion} successfully learn $p(z\vert\cond)$ with large-scale text-image datasets \cite{laion-dataset}. In other words, it understands the essential composition of natural images and the semantic relation between textual descriptions and images and can work as a great generative prior. Based on this finding, we propose our framework to leverage the power of pre-trained text-to-image latent diffusion models as a color prior for image colorization. 
\begin{figure}[h]
\centering
\includegraphics[width=1.0\linewidth]{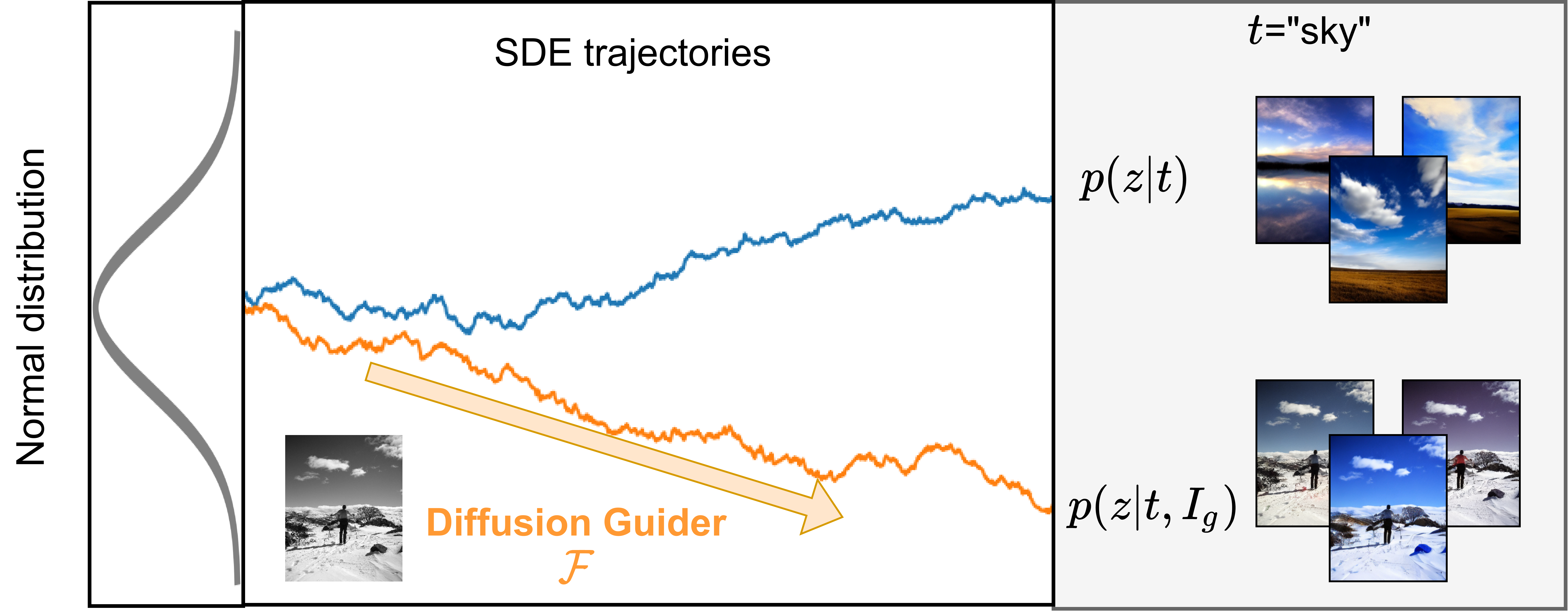}
   \caption{Given a grayscale input $I_g$, the diffusion guider ``hijacks" the SDE trajectories towards the target conditional data distribution $p(z|t,I_g)$.}
\label{fig:hijack-sde}
\end{figure}

Specifically, we propose our diffusion guider model $\mathcal{F}_{\theta}(\model, {I}_{g}, {I}_{h})$ that can guide the reverse diffusion process of a pre-trained diffusion model towards a certain subspace, in which $\texttt{rgb2gray}(\decoder(z_c)) \approx {I}_{g}$, as shown in Figure~\ref{fig:hijack-sde}. 
The Diffusion Guider share the a similar architecture of the denoising U-Net $\model(\circ, t, text)$ except it contains an additional encoder  $\mathcal{E}_g$ for the grayscale input image $I_g$ and the color hint point map $I_h$. As shown in Figure~\ref{fig:overview}, we take the feature map $F_\mathcal{F}$ from the output layer and each downsampling/upsampling layer of Diffusion Guider $\mathcal{F}$ and inject them into the corresponding layer of $\model$:
\begin{equation}
F_{{\model}_i}' = \texttt{Conv}_{1x1}(\texttt{Concat}(F_{{\model}_i}, \texttt{Conv}_{1x1}(F_{\mathcal{F}_i})))
\end{equation}

In this way, the Diffusion Guider takes the the grayscale image $I_g$ and a color hint point map $I_h$ as input and learns to modulate the inner features and the final output layers of $\model$ to generate a modified score, which leads the generation result conforming the grayscale image and optional user hints. In this way, we can consider $\mathcal{F}_{\theta}(\model, {I}_{g}, {I}_{h})$ as a large diffusion model conditioned on ${I}_{g}, {I}_{h}$.

\noindent\textbf{Loss function.}  
Treating $\mathcal{F}_{\theta}(\model, {I}_{g}, {I}_{h})$ as a conditional diffusion model, the loss function can be written as follows:
\begin{equation}
\resizebox{.9\hsize}{!}{$
    \lcolor = \expec_{\encoder(x), \epsilon \sim \mathcal{N}(0, 1),  t}\Big[ \Vert \epsilon - \mathcal{F}(\model, z_{t},t, text, I_g, I_h) \Vert_{2}^{2}\Big] \, $}
    \label{eq:colorloss}
\end{equation}
To reduce the learning difficulty and fully unitize diffusion prior, we freeze the weights of original diffusion model $\model$ during the training. 

\subsection{Lightness-aware VQVAE model}

\begin{figure}[h]
\centering
\input{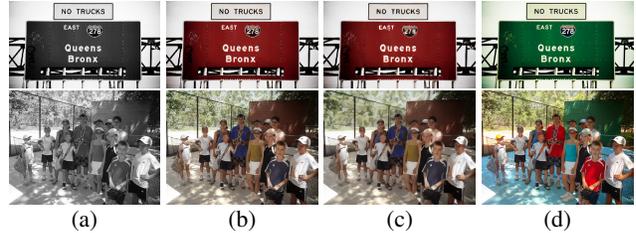}
   \caption{Ablation study on the lightness-aware VQVAE model. We show input images (a) and colorization results produced by our method without the lightness-aware VQVAE (b), our full method (c), and ground truth (d). Zoom in for better inspection.}
\label{fig:vae-apply}
\end{figure}

The output of the latent decoder $\decoder$ can contain serious distortions and artifacts as detailed structure and texture information is only approximated during the vector quantization process \cite{vq-vae,vq-gan}. In that case, the latent color prior $z_c$ is only a perceptually similar estimate for a color image $\tilde{I_c}=\decoder(z_c)$ whose intensity resembles $I_g$: $\texttt{rgb2gray}(\tilde{I_c}) \approx I_g$. This is unacceptable in our colorization task, as the structure and texture must be aligned well with the original grayscale input. 
Unfortunately, we have to remain in the latent space $\{z\}$ of the pre-trained latent diffusion model to leverage its power of feature learning and synthesis. Although using $\tilde{I_c}$ is not a viable option, due to the strong potential of the Stable Diffusion model, we can observe that the denoise diffused latent $z_c$ has already led to high-quality colorization to the input grayscale image, both globally and locally, as shown in Figure~\ref{fig:vae-apply}. Therefore, our primary concern now is the potential loss of information that occurs due to the VQ encoder. 

To address this issue, we propose a solution that fuses the user-guided latent diffused color prior $z_c$ with the pixel-level details from $I_g$, allowing for perfectly aligned colorization. To achieve this, we introduce the \ColorizationDecoder, which comprises a copy of the latent diffusion VQVAE decoder $\decoder$ and a grayscale encoder module $\GrayEncoder$. Rather than relying solely on the decoder $\decoder$ to recover the compressed structure and texture details, we design the grayscale encoder to shortcut and inject grayscale features from $I_g$ into the decoder. This provides the necessary hints for achieving pixel-perfect recovery during the decoding process.
Inspired by the idea of shortcut learning \cite{shortcut-learning}, the Grayscale Encoder $\GrayEncoder$ is almost identical to the original VQ-VAE encoder except we take the inner features $\{F_{\GrayEncoder_i}\}=\GrayEncoder(I_g)$ at each down-sampling layers as the output for the $\GrayEncoder$. After that, we add the processed grayscale features to the feature maps in the corresponding upsampling layers of the VQVAE decoder $\decoder$:
\begin{equation}
F_{\decoder_i}' = F_{\decoder_i} + \texttt{Conv}_{1x1}(F_{\GrayEncoder_i})
\end{equation}

To encourage \ColorizationDecoder~ to leverage the grayscale features and reduce the learning difficulty, we freeze the VQVAE decoder and optimize the Grayscale Encoder only during training. In other words, we treat $\texttt{Conv}_{1x1}(F_{\GrayEncoder_i})$ as learnable residuals \cite{resnet}. The architecture of \ColorizationDecoder~ resembles a UNet \cite{unet} disconnected in the middle layers. 

\noindent\textbf{Loss function.} We use a combination of $L_1$ loss, perceptual loss \cite{perceptual-loss}, and patch discriminator loss \cite{vq-gan} to encourage the quality of reconstruction. The loss function can be listed as follows:
\begin{equation}
L=\mathcal{L}_1+\lambda_{p}\mathcal{L}_p+\lambda_{d}\mathcal{L}_d
\end{equation}
where $\lambda_{p}=0.1$ and $\lambda_{d}$ is an adaptive weight following the strategy from \cite{vq-gan}.

\section{Experiments}
\begin{figure*}[!htb]
\centering
\includegraphics[width=1.0\linewidth]{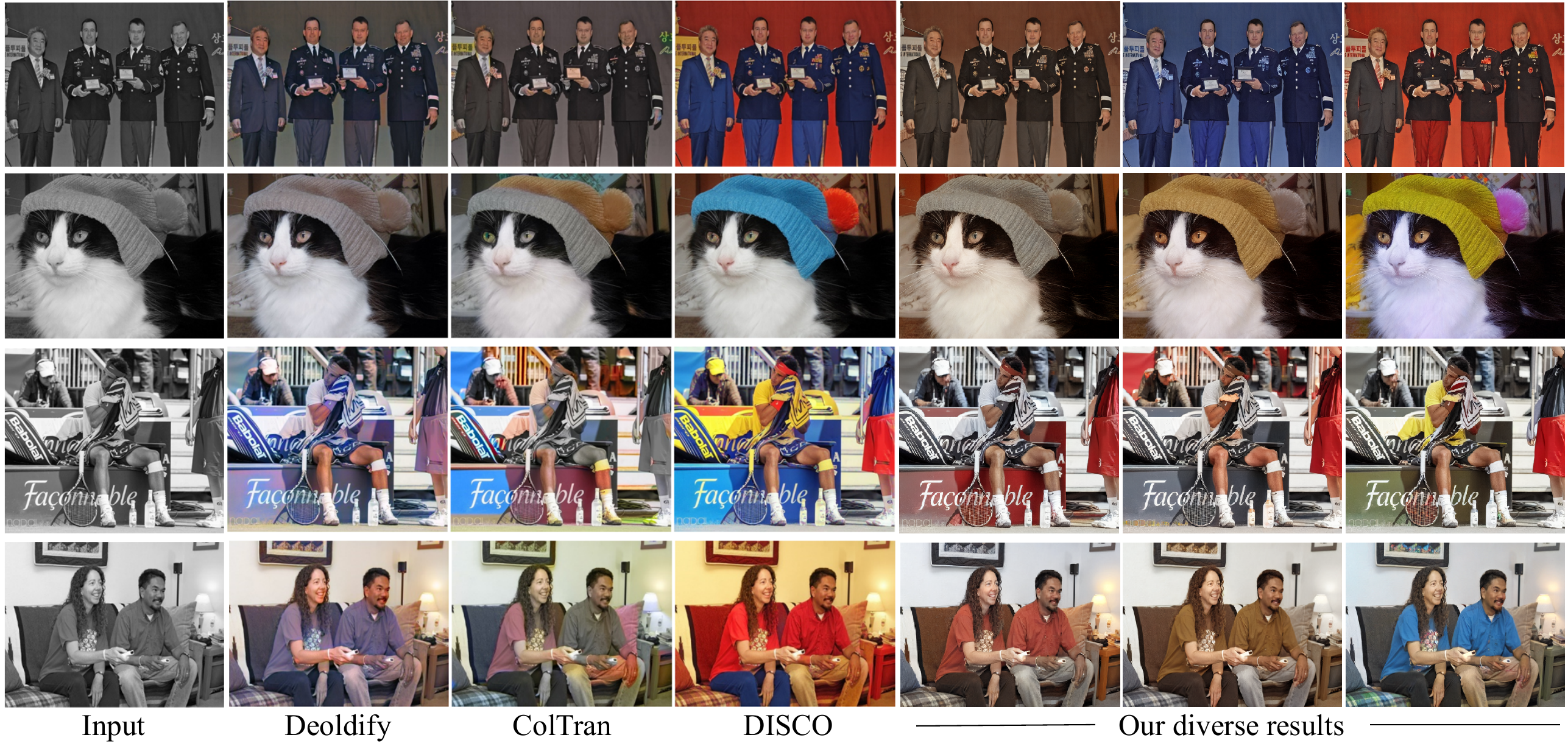}
\caption{Qualitative comparison with existing colorization methods.}
\label{fig:qualitative_eval}
\end{figure*}
\begin{figure*}[h]
\centering
\includegraphics[width=1.0\linewidth]{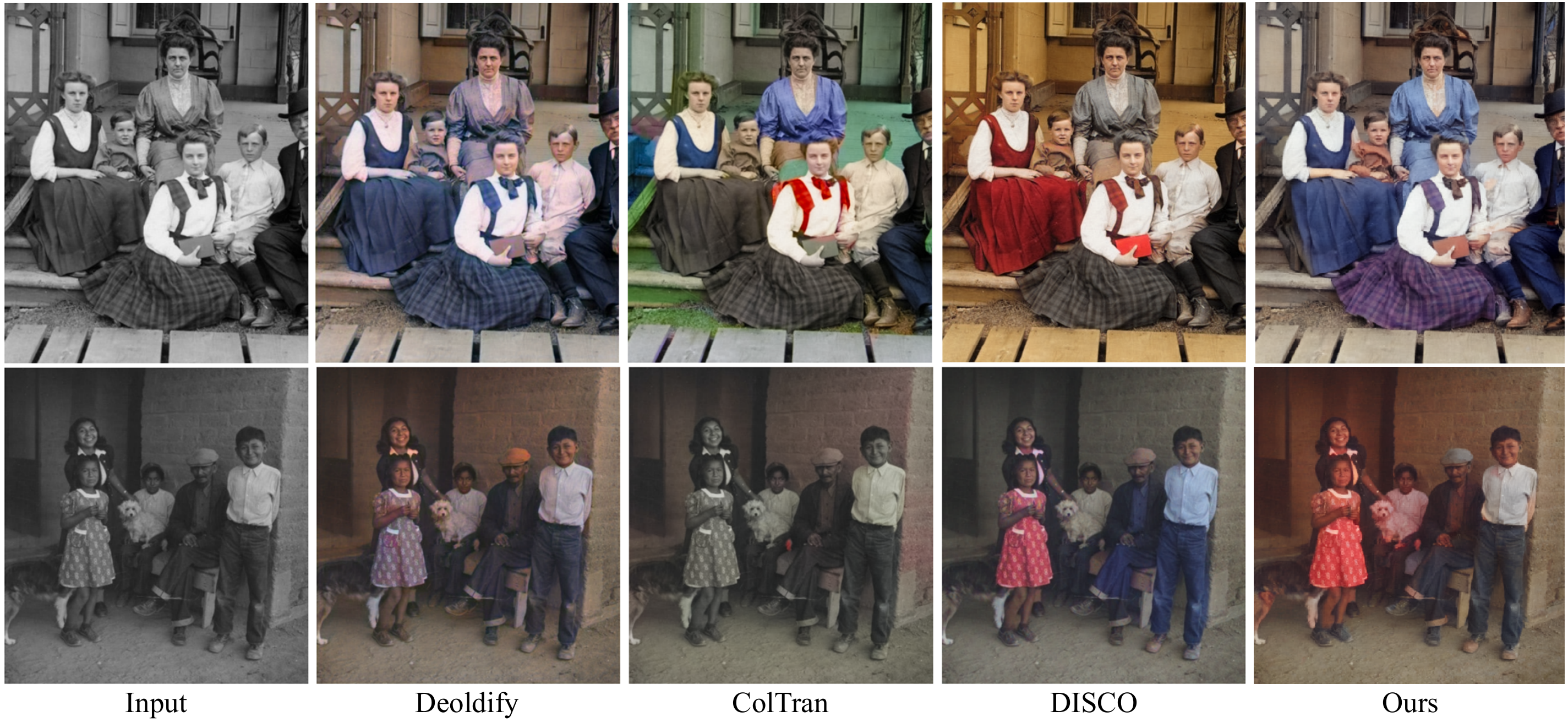}
\caption{Comparative showcase of colorizing historical photos (no ground truth available) from the KeystoneDepth dataset.}
\label{fig:history_eval}
\end{figure*}
\subsection{Implementation and training}

\noindent\textbf{Implementation details}. We implement our framework in PyTorch 1.13 based on the open-source codebase of latent diffusion \cite{latent-diffusion}. We use pre-trained \texttt{miniSD} \cite{mini-sd}, a fine-tuned Stable Diffusion v1.4 model \cite{latent-diffusion} on image size 256x256 as the text-to-image latent diffusion prior inside our framework.   

\noindent\textbf{Training setup}. We train the \PriorModel~on the COCO dataset \cite{mscoco} and the \ColorizationDecoder~on the ImageNet dataset \cite{imagenet}. We use the AdamW optimizer \cite{adamw} with learning rate $lr=0.00001$ for both models. 
For generating the color hint map in \PriorModel, we employed SLIC superpixel \cite{achanta2012slic} and color quantization methods. During training, we randomly selected $30\%-50\%$ of the color hint regions from the quantized superpixel map of the original image. To enable unconditional colorization, we randomly replaced the original caption with dummy captions such as \textit{""}, \textit{"color photo"}, and \textit{"high quality photo"} at a chance of 50 percent during the training.

\begin{table*}[!ht]
        \setlength\tabcolsep{3pt}
 \centering
 \resizebox{1\textwidth}{!}{
 
	\begin{tabular}{lc|ccccc|ccccc}
		\toprule
		\multirow{2}*{Method}& \multirow{2}*{Venue} & \multicolumn{5}{c|}{ImageNet (10k)} &\multicolumn{5}{c}{COCO-Stuff (5k)}  \\
		~&~ &FID $\downarrow$ &Colorfulness $\uparrow$ &PSNR $\uparrow$ &SSIM $\uparrow$ &LPIPS $\downarrow$  &FID $\downarrow$ &Colorfulness $\uparrow$ &PSNR $\uparrow$ &SSIM $\uparrow$ &LPIPS $\downarrow$ \\
		\midrule
		CIColor \cite{zhang2016colorful}     & ECCV'16 &11.58 &42.89 &21.96 &0.897 &0.224  &21.44 &42.46 &22.08 &0.902 &0.217  \\
		UGColor \cite{zhang2017real}         & TOG'17 &6.85 &27.91 &\textbf{24.26} &\textbf{0.919} &\textbf{0.174}  &14.74 &28.64 &\textbf{24.34} &\textbf{0.924} &\textbf{0.165}\\
		Deoldify \cite{antic2019github}      & -'19 &5.78 &23.41 &23.34 &0.907 &0.188  &12.75 &23.62 &23.49 &0.914 &0.181  \\
		InstColor \cite{su2020instance}      & CVPR'20 &7.35 &25.54 &22.03 &0.909 &0.919  &12.24 &29.38 &22.35 &0.838 &0.238  \\
		ChromaGAN \cite{vitoria2020chromagan}& WACV'20 &9.60 &29.34 &22.85 &0.876 &0.230  &20.57 &29.34 &22.74 &0.871 &0.233  \\
		ColTran \cite{kumar2021colorization} & ICLR'21 &6.37 &38.64 &21.81  &0.892  &0.218  &11.65 &38.95 &22.11 &0.898 &0.210  \\
		DISCO \cite{disco-color}             &TOG'22 & \textbf{5.57} & \textbf{51.43} &20.72 &0.862 &0.229  &{10.59} &\textbf{52.85} &20.46 &0.851 &0.236  \\
            Ours                                 &-& {6.11} & 32.18 & 21.64 &  0.861 & 0.254 & \textbf{8.23} & 31.32 & 22.12  & 0.860  &  0.239 \\
            \hline
            Ours\textsuperscript{\textdagger}                                 &-& 4.21 & 46.27 & 20.88 &  0.845 & 0.251 & 7.66 & 40.29 & 21.35  & 0.849  &  0.247 \\
            Ours\textsuperscript{\textdaggerdbl}                                 &-& 6.37 & 41.62 & 20.66 &  0.847 & 0.269 & 8.66 & 42.91 & 20.97  & 0.843  &  0.256 \\
		\bottomrule
	\end{tabular}
 
 }
	\caption{Quantitative results on the validation sets from different methods. The best results are in \textbf{bold} font. Ours\textsuperscript{\textdagger} uses generated captions from BLIP captioning model \cite{li2022blip} as additional inputs. Ours\textsuperscript{\textdaggerdbl} uses prompt \textit{"a high quality colorful photo"} and negative prompt \textit{"color bleeding"} as additional inputs. }
	\label{tab:quanti_eval}
\end{table*}

\subsection{Comparisons}
\noindent\textbf{Dataset.} We evaluate our method and existing representative works on standard evaluation benchmarks, \ie, ImageNet~\cite{imagenet} and COCO-Stuff~\cite{caesar2018coco}. Following the prior protocols~\cite{su2020instance,zhang2016colorful}, we evaluated all methods in ImageNet-ctest (10k images), which is a subset of the ImageNet validation set. For COCO-Stuff, we adopt its validation set (5k images) to perform evaluations. 

\noindent\textbf{Metrics.} The main evaluation aspects of image colorization are perceptual realism and color vividness. Therefore, we adopt Fréchet Inception Score (FID)~\cite{heusel2017gans} to measure the distribution similarity between the prediction and the ground truth, which reflects perceptual realism to some extent. To assess color vividness, we employed the Colorfulness metric\cite{hasler2003measuring}, which is similar to human vision perception. Additionally, we also reported evaluation results using PSNR, SSIM, and LPIPS~\cite{zhang2018unreasonable}, as previously done in works such as \cite{su2020instance,vitoria2020chromagan,zhang2017real,disco-color}. However, it is important to note that plausible colorization results may have significant color differences from the ground truth, and hence these metrics may not accurately reflect the actual performance, but only serve as a reference.

\noindent\textbf{Unconditional colorization.} For unconditional colorization, we compare with three types of state-of-the-art methods that can colorize grayscale images without user hints: a) CNN-based method: CIColor~\cite{zhang2016colorful}, UGColor~\cite{zhang2017real}, Deoldify~\cite{antic2019github}, InstColor~\cite{su2020instance}, ChromaGAN~\cite{vitoria2020chromagan}, b) Transformer-based method: ColTran~\cite{kumar2021colorization}, and c) Hybrid structure with CNN and Transformer: DISCO~\cite{disco-color}. 
We show the qualitative comparison in Figure~\ref{fig:qualitative_eval} and report the quantitative results in Table~\ref{tab:quanti_eval}. Our proposed method achieved state-of-the-art performance in both evaluations. Additionally, due to the stochastic nature of the reverse diffusion process, our framework can natively generate diverse results. 

\noindent\textbf{Conditional colorization. } 
\begin{figure}[h]
\centering
\includegraphics[width=1.0\linewidth]{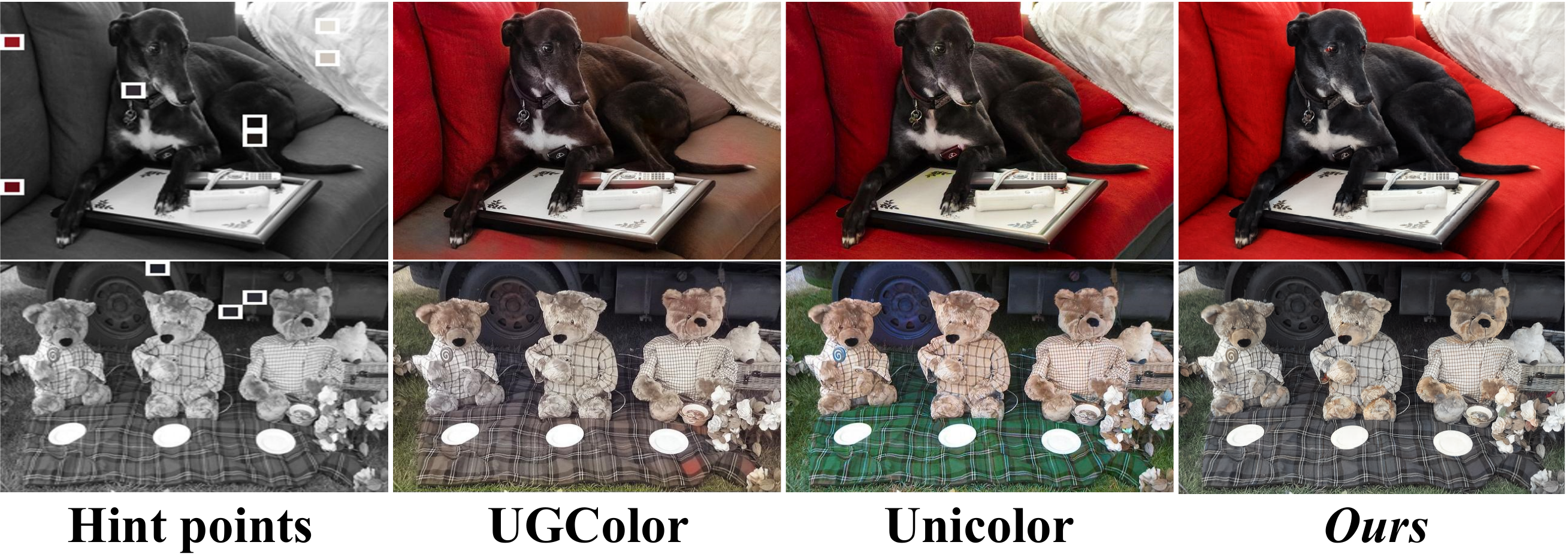}
   \caption{Visual comparisons with the hint point-based method (UGColor~\cite{zhang2017real} and UniColor~\cite{unicolor}).}
\label{fig:hint-point-based}
\end{figure}
For hint point-based colorization, we compare our framework with two recent state-of-the-art methods: a CNN-based user-guided method\cite{zhang2017real} and a transformer-based multi-modal colorization method\cite{unicolor}. As shown in Figure~\ref{fig:hint-point-based}, our method propagates the hint point in the corresponding consistently and smoothly, while \cite{zhang2017real} and \cite{unicolor} suffer from color bleeding problems. For the regions without user hints, our method can also generate harmonized results with the user-guided regions. 

\begin{figure}[h]
\centering
\includegraphics[width=\linewidth]{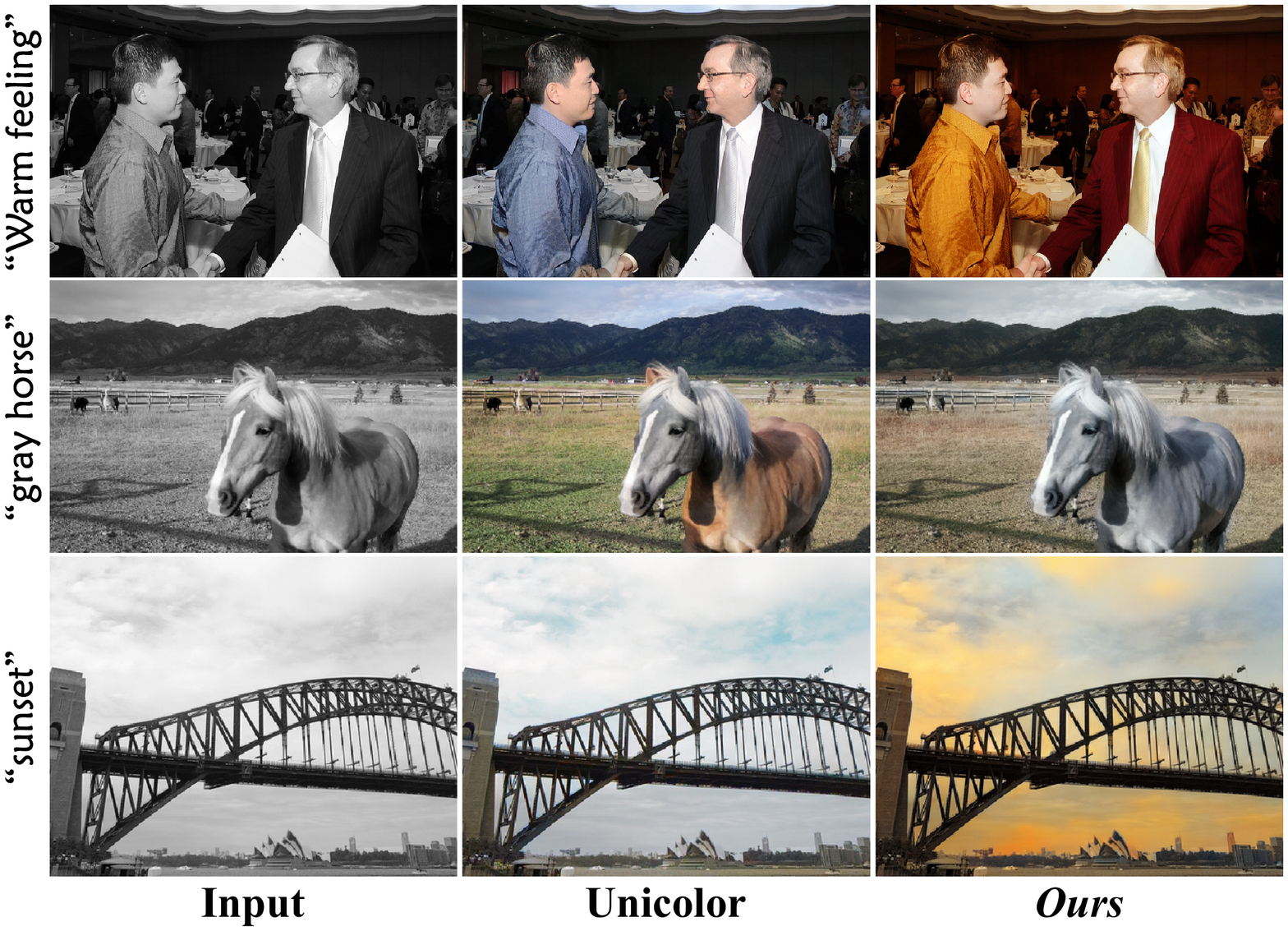}
\centering
   \caption{Visual comparisons with the text-based method (UniColor~\cite{unicolor}) sampled from COCO-Stuff.}
\label{fig:text-based}
\end{figure}

For text-based colorization, we compare with UniColor using the prompt with and without explicit color descriptions. As shown in Figure~\ref{fig:text-based}, our method can understand prompts like \textit{sunsetting} and \textit{warm feeling}, while UniColor fail to generate semantic-aligned colorization results. The existing methods mainly focus on object-color pairs and therefore cannot work for high-level color descriptions. 
\subsection{User study}

\begin{figure}[h]
    \centering
\includegraphics[width=0.45\textwidth]{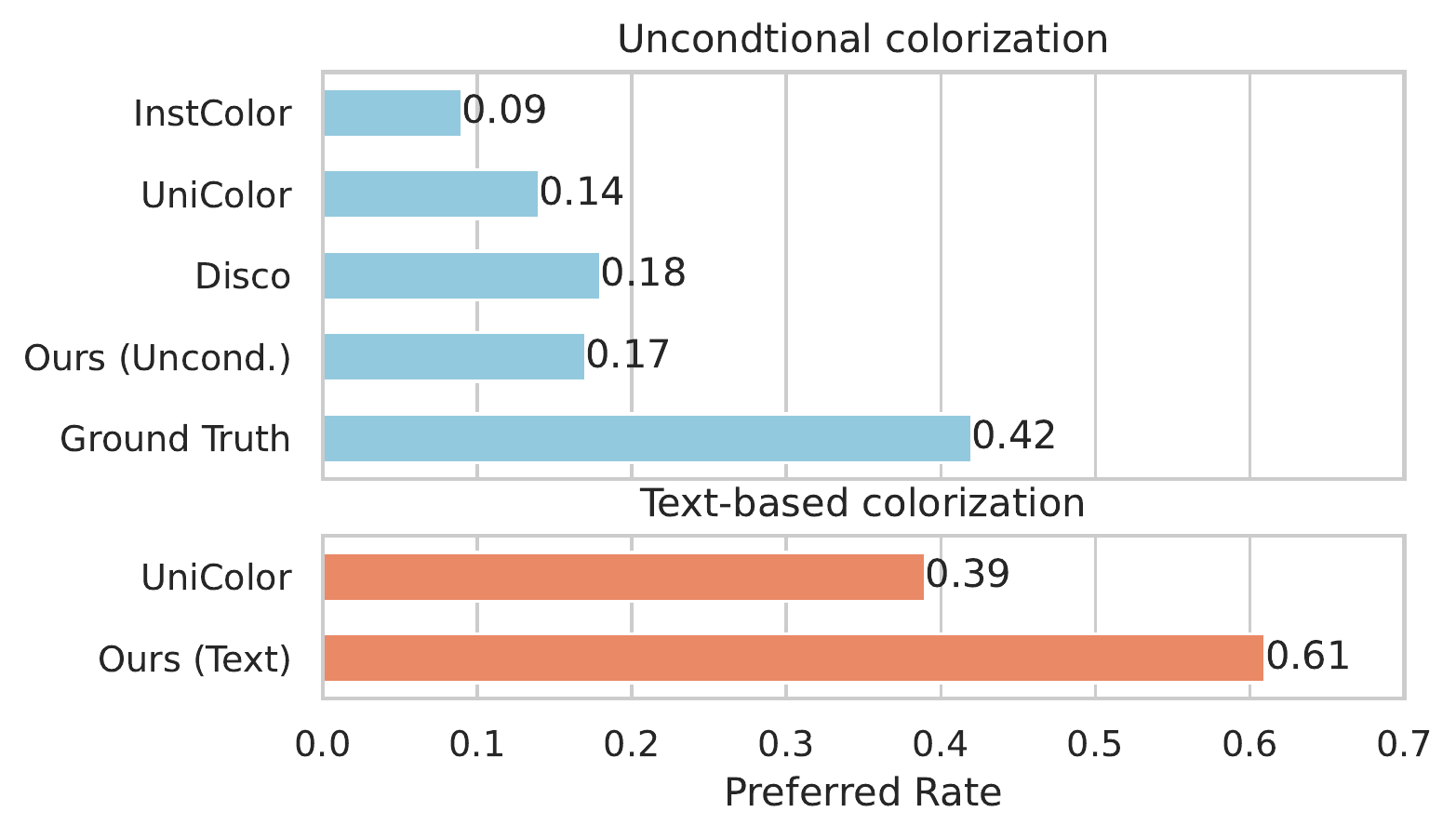}
\caption{Results of user study on unconditional and text-based colorization.}
\label{fig:user-study}
\end{figure}

To further verify the effectiveness of our method, we conducted a user study and present the results in this subsection. We design two testing scenarios: unconditional and text-based colorization through the user study. 
For each scenario, we randomly sampled 30 images from the COCO-Stuff validation dataset \cite{mscoco}. Each participant was ask to complete 15 questions per scenario. For each question, the participants were shown a series of color images, including the colorized images by different methods and the ground truth image (only shown for unconditional colorization), placed in random order. Participants were then asked to select the best colorization result based on perceptual realism and consistency with the input prompt (if applicable). In total, 67 participants with good vision and color recognition successfully completed the evaluation. We show the result in Figure~\ref{fig:user-study}. For unconditional colorization, all methods still lag behind the ground truth images. Our method achieved similar perceptual realism with Disco \cite{disco-color} and outperformed the other methods. For text-based colorization, our method outperformed UniColor \cite{unicolor} in terms of perceptual realism and consistency with the input prompt.

\subsection{Ablation study}
\begin{table}[h]


\centering
\resizebox{0.48\textwidth}{!}{
\begin{tabular}{lccccc}
  \toprule
  Model& Train Steps & R-FID $\downarrow$ & PSNR $\uparrow$ & SSIM $\uparrow$ & PSIM $\downarrow$\\
  \midrule
  \texttt{kl-f8}   &  246k    &  4.99& 23.4                      & 0.69 & 1.01 \\
  \texttt{ft-EMA}   &  313k      & 4.42 & 23.8                      & 0.69 & 0.96  \\
  \texttt{ft-MSE}    &  280k     & 4.70 &  24.5                     & 0.71 & 0.92 \\
  Ours             &  44k &\textbf{1.88}    &  \textbf{35.3}  & \textbf{0.97}  & \textbf{0.03} \\
  \bottomrule
  \end{tabular}
  }

\caption{
  Quantitative comparisons of reconstructions by different decoders on the COCO-Stuff validation set ($256^2$).
}
\label{tab:decoder}

\end{table}

\begin{figure}[h]
\centering
\includegraphics[width=0.9\linewidth]{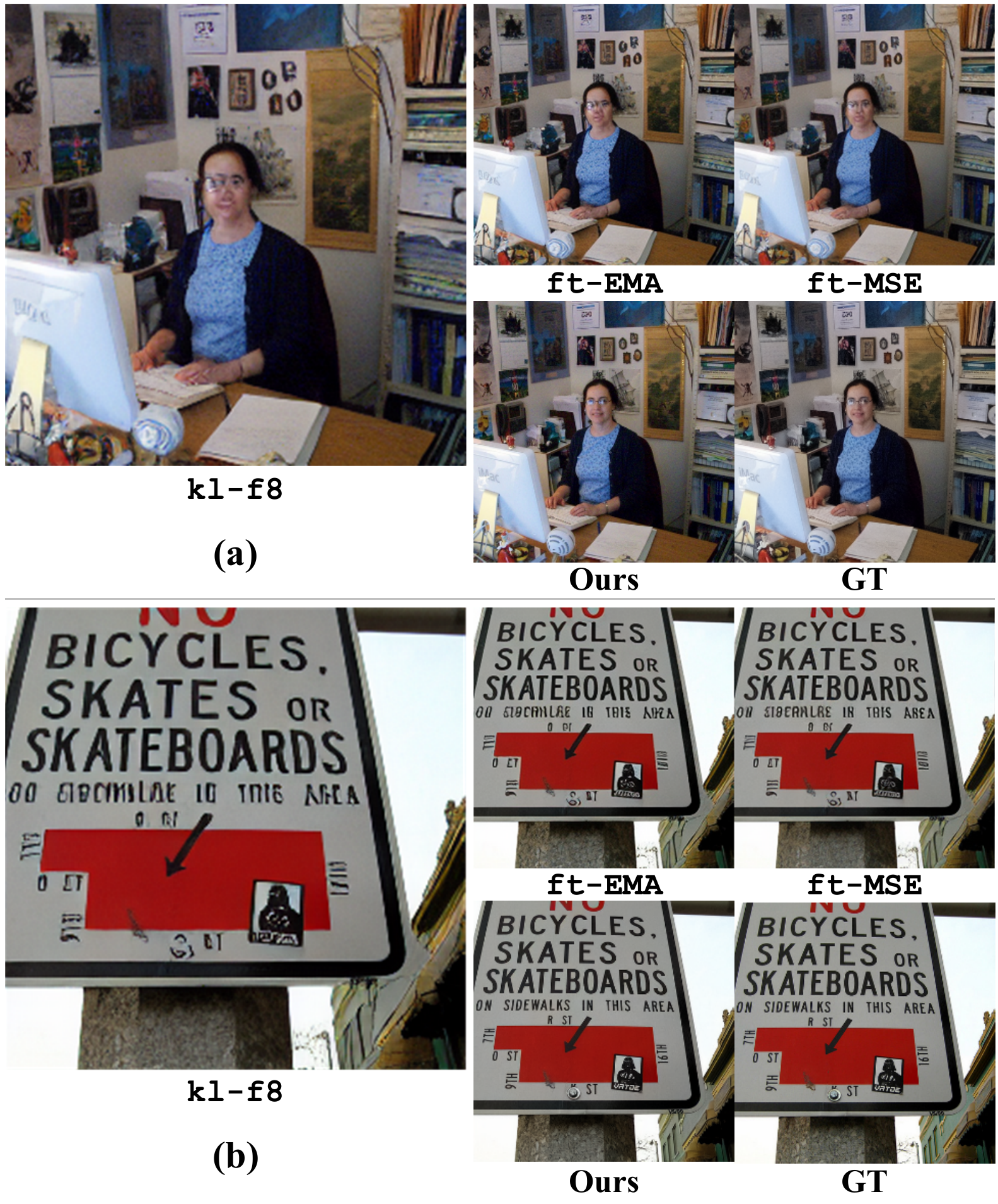}
   \caption{Visual comparison of reconstructions with different decoders sampled from the COCO-Stuff validation set.}
\label{fig:vae-out}
\end{figure}

\begin{figure*}[h]
    \centering
    \input{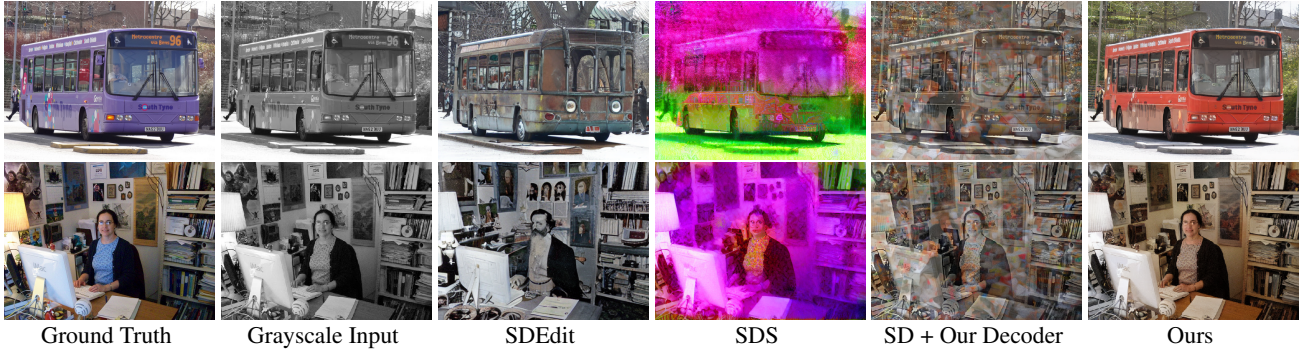}
\caption{Visual comparison with different diffusion-based image editing methods.}
\label{fig:editing}
\end{figure*}

\textbf{The effect of \ColorizationDecoder.}
To verify whether our \ColorizationDecoder~ can better reconstruct the images with the grayscale features, we compare \ColorizationDecoder~with the original \texttt{kl-f8} VQVAE decoder used by Stable Diffusion and two fine-tuned versions of the VQVAE decoder: \texttt{ft-EMA} and \texttt{ft-MSE} \cite{sd-vae}. As shown in Table~\ref{tab:decoder}, our \ColorizationDecoder~ achieves the best performance with the help of grayscale features. The visual comparison is shown in Figure~\ref{fig:vae-out}. Our \ColorizationDecoder~can produce the sharpest results. 


\textbf{The effect of \PriorModel.}
The \PriorModel~ learns to guide the denoising diffusion process towards a certain subspace in which the latent conforms to the input grayscale image. In this section, we first create a baseline without using the proposed guider model $\guider$ (e.g., vanilla latent diffusion with the proposed \ColorizationDecoder). We then create the other baseline using different guidance mechanisms: SDEdit \cite{sdedit}, and Score Distillation Sampling (SDS) \cite{dreamfusion}. We show the results in Figure~\ref{fig:editing}. For evaluation, we use the default guidance scale and DDIM sampler \cite{ddim} with 50 sampling steps. For SDEdit, we set the sampling strength to $0.45$. The vanilla-diffusion and SDEdit based sampling approach deviate from the grayscale inputs. The SDS method cannot produce visually satisfying results due to the gradient propagation through the VQVAE. Our method outperforms the mentioned baselines. 

\subsection{Limitations}

\noindent\textbf{Spatial resolution.} Our framework's performance may degrade when processing input images with spatial dimensions less than 256x256, due to the usage of text-to-image latent diffusion prior \cite{latent-diffusion}. It is necessary for users to upscale smaller images before using our framework. 
\begin{figure}[h]
    \centering
    \input{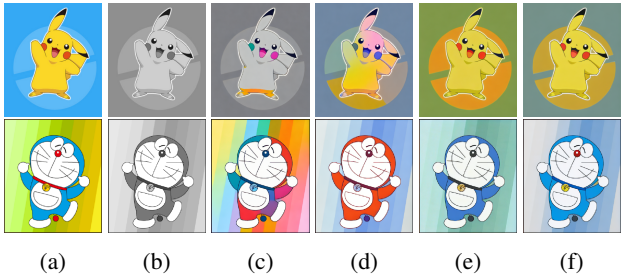}
\caption{(a) Ground truth; (b) Input; (c-d) Unconditional colorization; (e1) \textit{"pokemon"}; (e2) \textit{"japanese cartoon"}; (f1) \textit{"pikachu"}; (f2) \textit{"doraemon"}. Our framework might fail for unconditional colorization of non-natural images due to the domain gaps introduced in the training. 
}
\label{fig:domain-gap}
\end{figure}

\noindent\textbf{Generalization.} Despite the ability of stable diffusion to generate non-natural images, our current framework may not effectively generalize to such domains due to its training on natural image datasets, as shown in Figure~\ref{fig:domain-gap}. We posit that this domain gap is a consequence of our training strategy rather than a limitation of our framework's capability. To address this issue, training on datasets that correspond to the target domain may prove helpful in mitigating the domain gap. Furthermore, users can incorporate explicit textual hints related to the input image to alleviate the domain gap. 

\noindent\textbf{Runtime performance.} The diffusion model utilized in our framework requires an time-consuming iterative sampling process, which may take minutes to complete on CPU devices. Hence, it is imperative that users should utilize GPUs when running our pipeline. Despite recent studies on efficient diffusion samplers that have somehow alleviate this issue, the performance of our framework remains limited on CPU or low-end GPU devices.

\noindent\textbf{Ethical concerns.} We have leveraged pre-trained models based on Stable Diffusion v1.4 \cite{latent-diffusion} and miniSD \cite{mini-sd}, which were trained on the LAION dataset \cite{laion-dataset}, a dataset known to have social and cultural biases. It remains unclear how these biases can manifest themselves in the color space and potentially influence the outputs of our framework.

\section{Conclusion}
In this paper, we propose an effective colorization pipeline piggybacking on the pre-trained latent diffusion models for realistic and diverse image colorization. 
To reuse the color prior knowledge in the powerful T2I model, we design a diffusion guider model to guide the pre-trained latent diffusion model to output a latent color prior that conforms to the visual semantics of the grayscale input. A novel lightness-aware VQVAE model is proposed for pixel-perfect aligned colorization against the input by shortcuting the grayscale information to the decoder. In addition, our method can also accept additional user controls for conditional colorization. 
As evidenced by extensive evaluation, our method achieves  state-of-the-art performance on both unconditional and conditional colorization.

{\small
\bibliographystyle{ieee_fullname}
\bibliography{egbib}
}
\clearpage
\onecolumn
\appendix

\section{Showcase for text-based image colorization}
\begin{figure*}[ht]
    \centering
    \captionsetup[subfigure]{labelformat=empty}
\def\myhighreswidth{0.22}
\def\myhighresoffset{0.0}
\centering
\begin{tabular}[c]{cccc}
\multirow{2}{*}{
\begin{subfigure}[c]{\myhighreswidth\textwidth}\includegraphics[width=\textwidth]{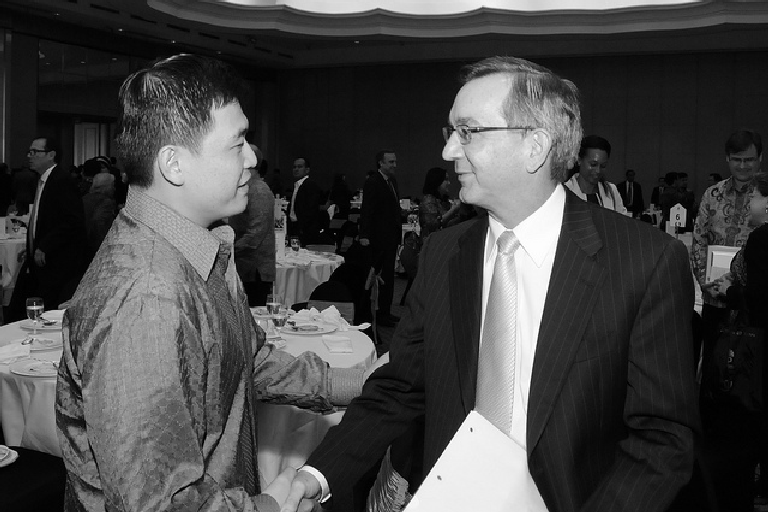}
        \subcaption{Grayscale Input}
        \end{subfigure}\hspace{\myhighresoffset\textwidth}
        }
        &
        \begin{subfigure}[c]{\myhighreswidth\textwidth}\includegraphics[width=\textwidth]{graphs/text_showcase/warm_feeling_gt.jpg}
        \subcaption{Ground Truth}
        \end{subfigure}\hspace{\myhighresoffset\textwidth}
        &
        \begin{subfigure}[c]{\myhighreswidth\textwidth}\includegraphics[width=\textwidth]{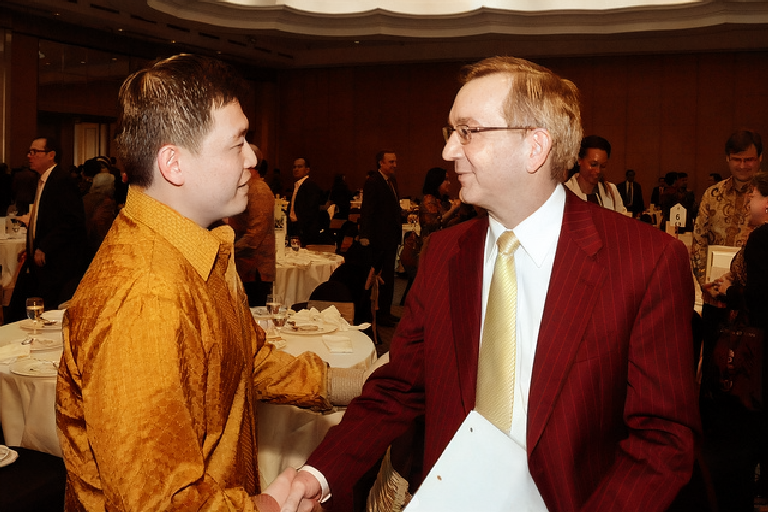}
        \subcaption{"warm feeling"}
        \end{subfigure}\hspace{\myhighresoffset\textwidth}
        &
        \begin{subfigure}[c]{\myhighreswidth\textwidth}\includegraphics[width=\textwidth]{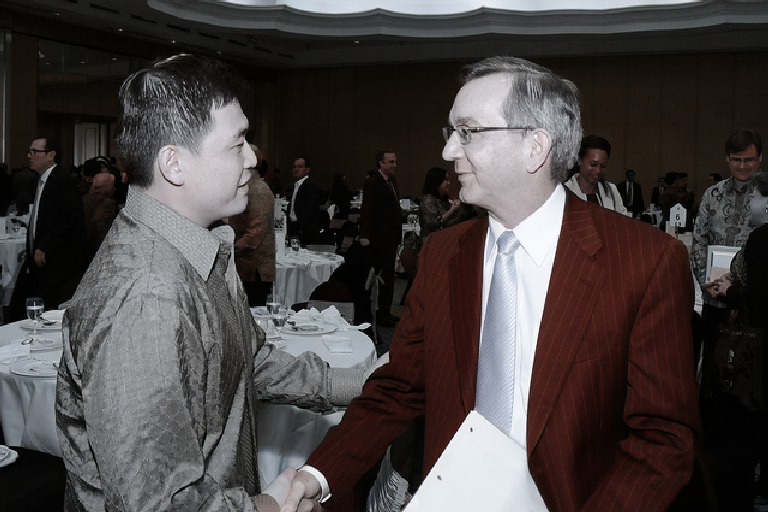}
        \subcaption{"cold feeling"}
        \end{subfigure}\hspace{\myhighresoffset\textwidth}
        \\
        &
        \begin{subfigure}[c]{\myhighreswidth\textwidth}\includegraphics[width=\textwidth]{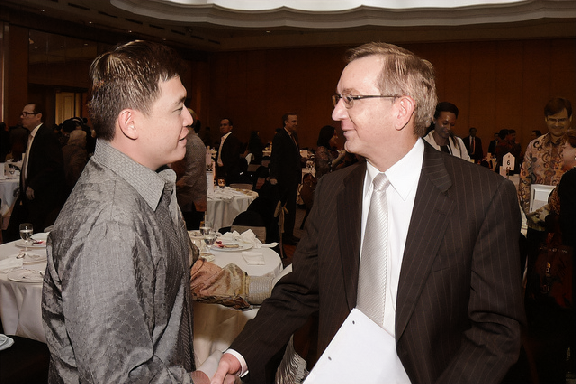}
        \subcaption{Unconditional colorization}
        \end{subfigure}\hspace{\myhighresoffset\textwidth}
        &
        \begin{subfigure}[c]{\myhighreswidth\textwidth}\includegraphics[width=\textwidth]{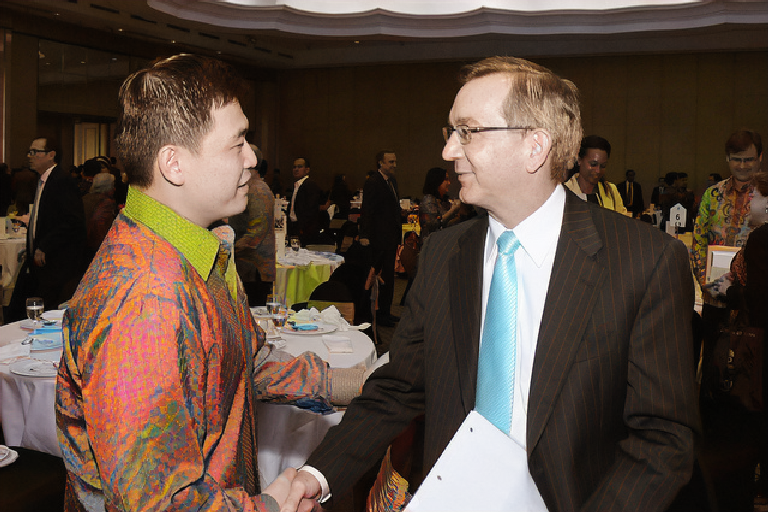}
        \subcaption{"colorful"}
        \end{subfigure}
        &
        \begin{subfigure}[c]{\myhighreswidth\textwidth}\includegraphics[width=\textwidth]{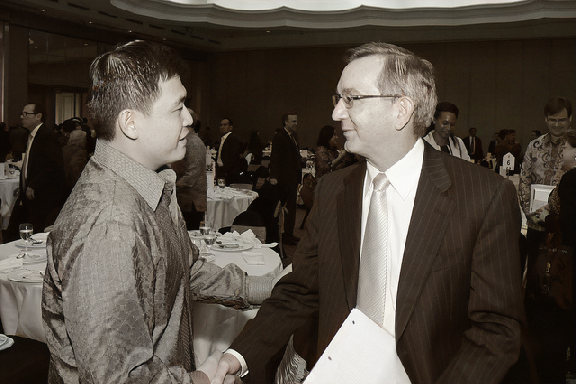}
        \subcaption{"old photo"}
        \end{subfigure}\hspace{\myhighresoffset\textwidth}
\\
\multirow{2}{*}{
\begin{subfigure}[c]{\myhighreswidth\textwidth}\includegraphics[width=\textwidth]{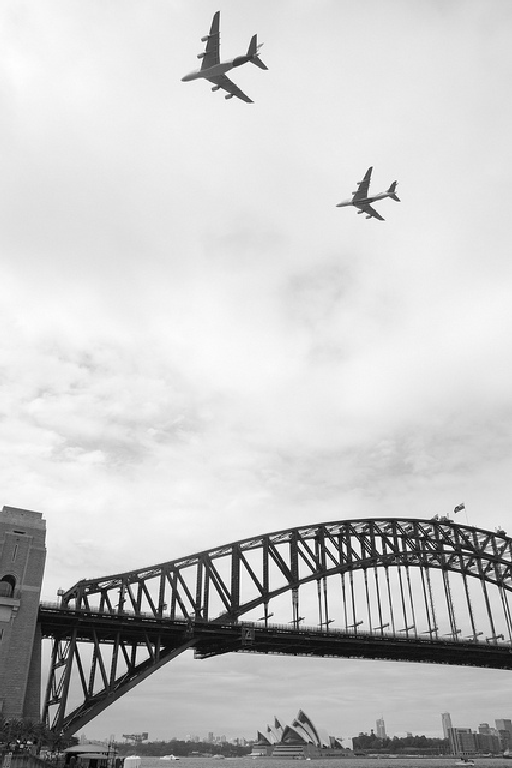}
        \subcaption{Grayscale Input}
        \end{subfigure}\hspace{\myhighresoffset\textwidth}
        }

        &
        \begin{subfigure}[c]{\myhighreswidth\textwidth}\includegraphics[width=\textwidth]{graphs/text_showcase/01gt.jpg}
        \subcaption{Ground Truth}
        \end{subfigure}\hspace{\myhighresoffset\textwidth}
        &
        \begin{subfigure}[c]{\myhighreswidth\textwidth}\includegraphics[width=\textwidth]{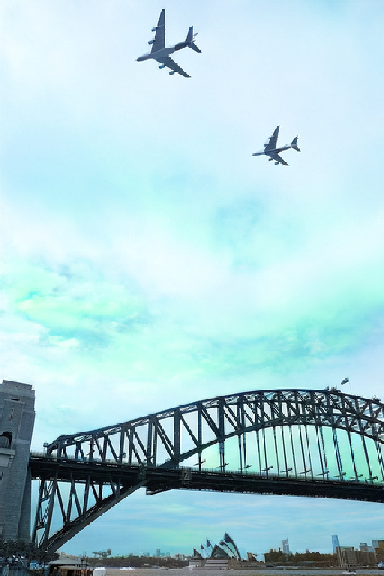}
        \subcaption{"aurora"}
        \end{subfigure}\hspace{\myhighresoffset\textwidth}
        &
        \begin{subfigure}[c]{\myhighreswidth\textwidth}\includegraphics[width=\textwidth]{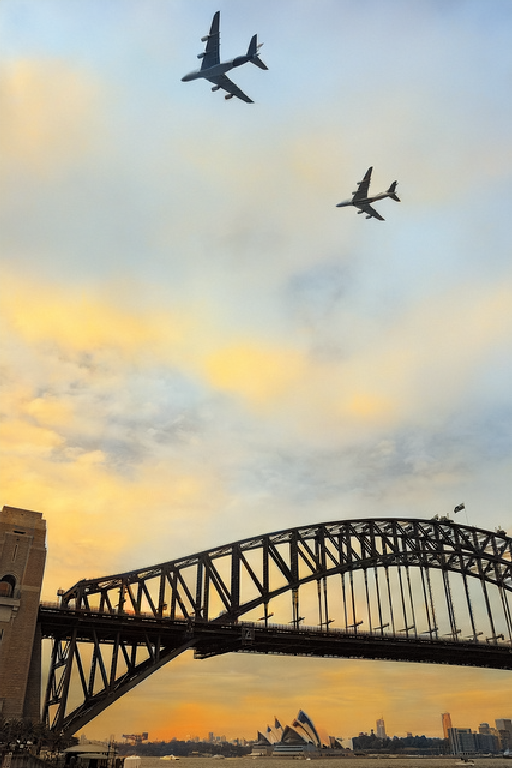}
        \subcaption{"sunset"}
        \end{subfigure}\hspace{\myhighresoffset\textwidth}
        \\
        &
        \begin{subfigure}[c]{\myhighreswidth\textwidth}\includegraphics[width=\textwidth]{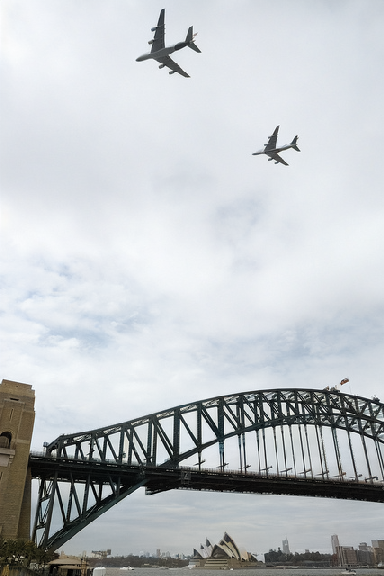}
        \subcaption{Unconditional colorization}
        \end{subfigure}\hspace{\myhighresoffset\textwidth}
        &
        \begin{subfigure}[c]{\myhighreswidth\textwidth}\includegraphics[width=\textwidth]{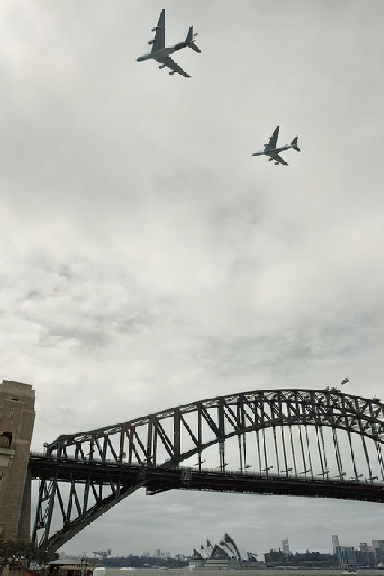}
        \subcaption{"rainy day"}
        \end{subfigure}\hspace{\myhighresoffset\textwidth}
        &
        \begin{subfigure}[c]{\myhighreswidth\textwidth}\includegraphics[width=\textwidth]{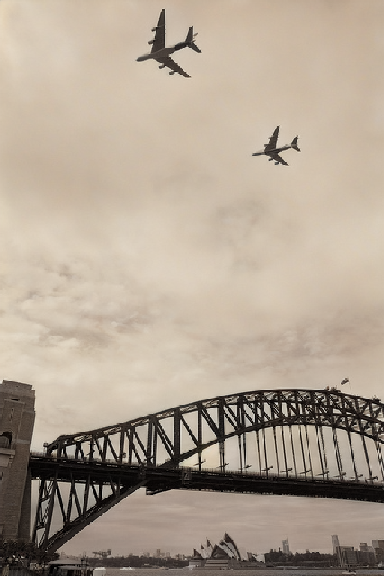}
        \subcaption{"at night"}
        \end{subfigure}

\end{tabular}
\caption{Colorization results with various text inputs.}
\label{fig:text_showcase}
\end{figure*}

In contrast to existing text-based image colorization methods such as \cite{text-guided-color, text2color, unicolor, l-code}, our framework supports arbitrary text inputs. To demonstrate this capability, we present a set of colorization results with various text inputs in Figure~\ref{fig:text_showcase}. 

\section{Additional comparative results}
\begin{figure*}[ht]
    \centering
    \input{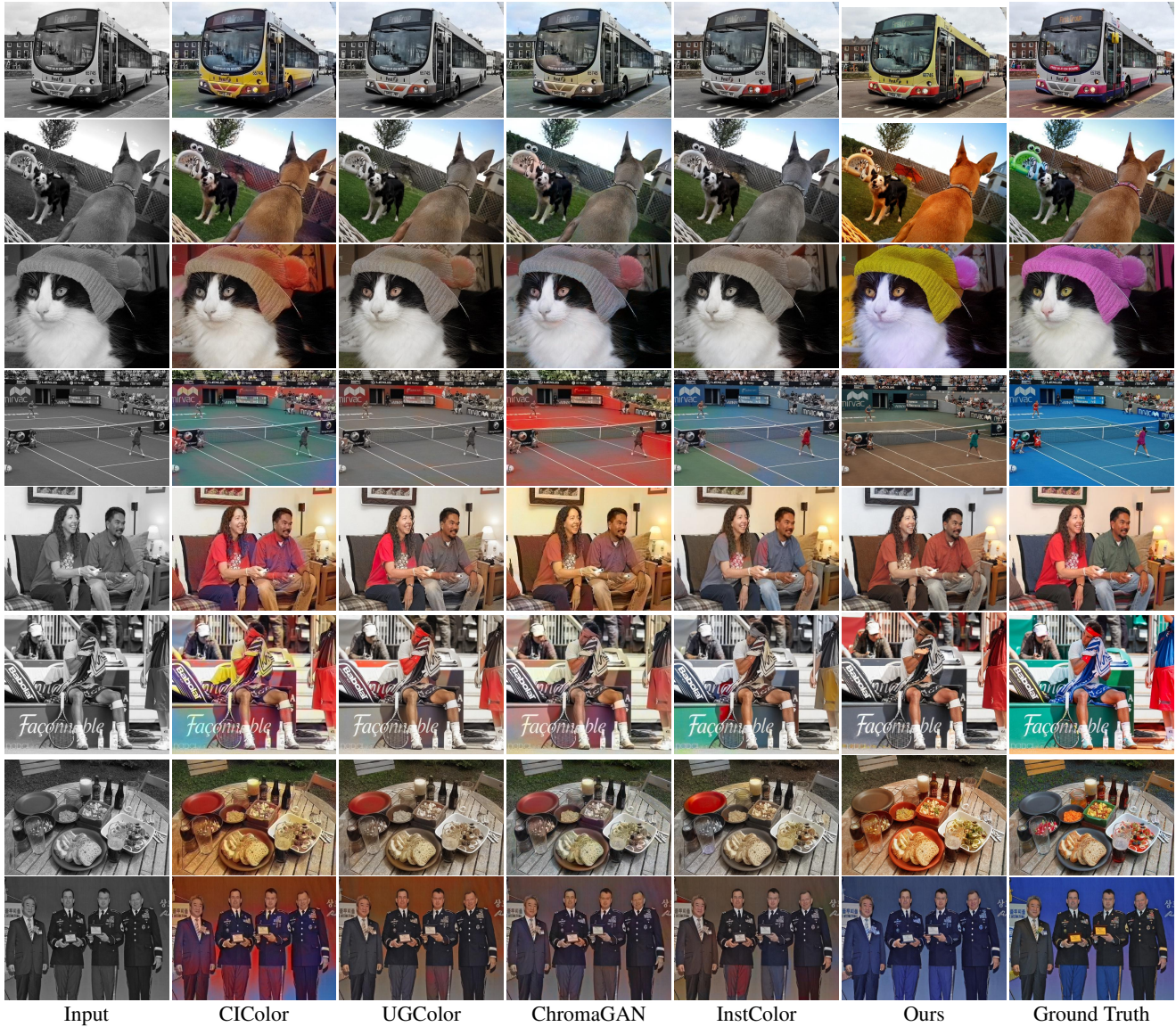}
\caption{Comparative showcase of challenging cases for unconditional image colorization.}
\label{fig:uncond0}
\end{figure*}

In the main paper, we only present a qualitative comparison with three of the most competitive colorization methods: Deoldify \cite{antic2019github}, ColTran \cite{kumar2021colorization}, and Disco \cite{disco-color}. In this section, we provide a comprehensive visual comparison with recent colorization methods, including CIColor \cite{zhang2016colorful}, UGColor \cite{zhang2017real}, Deoldify \cite{antic2019github}, ChromaGAN \cite{vitoria2020chromagan}, InstColor \cite{su2020instance}, ColTran \cite{kumar2021colorization}, UniColor \cite{unicolor}, and Disco \cite{disco-color}. The results are shown in Figure~\ref{fig:uncond0}, Figure~\ref{fig:uncond1}, Figure~\ref{fig:uncond2}, and Figure~\ref{fig:uncond3}. 

\begin{figure*}[ht]
    \centering
    \input{graphs/uncond1}
\caption{Comparative showcase of challenging cases for unconditional image colorization.}
\label{fig:uncond1}
\end{figure*}

\begin{figure*}[ht]
    \centering
    \input{graphs/uncond2}
\caption{Comparative showcase for old photo image colorization.}
\label{fig:uncond2}
\end{figure*}

\begin{figure*}[ht]
    \centering
    \input{graphs/uncond3}
\caption{Comparative showcase for old photo image colorization.}
\label{fig:uncond3}
\end{figure*}

\begin{figure*}[ht]
    \centering
    \input{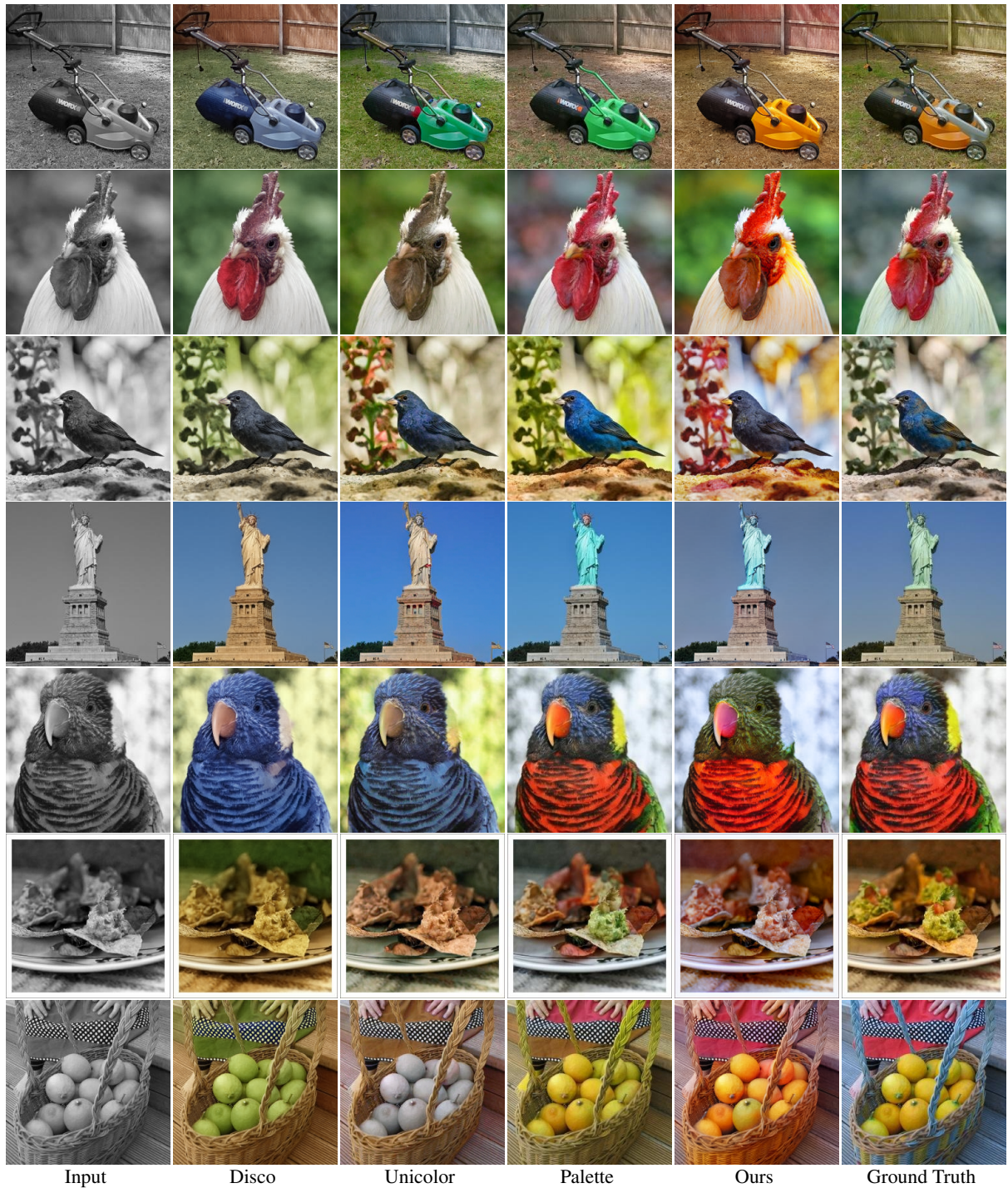}
\caption{Visual comparison with Palette diffusion model.}
\label{fig:palette}
\end{figure*}

We also present a visual comparison with Palette \cite{palette-diffusion}, which is currently the state-of-the-art diffusion-based method for image colorization. However, it should be noted that the source code and pre-trained model of Palette are not publicly available. Therefore, we use the images provided in the original paper for the comparison. The results are shown in Figure~\ref{fig:palette}. 

\section{Network architecture}
\begin{figure}[h]
    \centering
\includegraphics[width=1.0\textwidth]{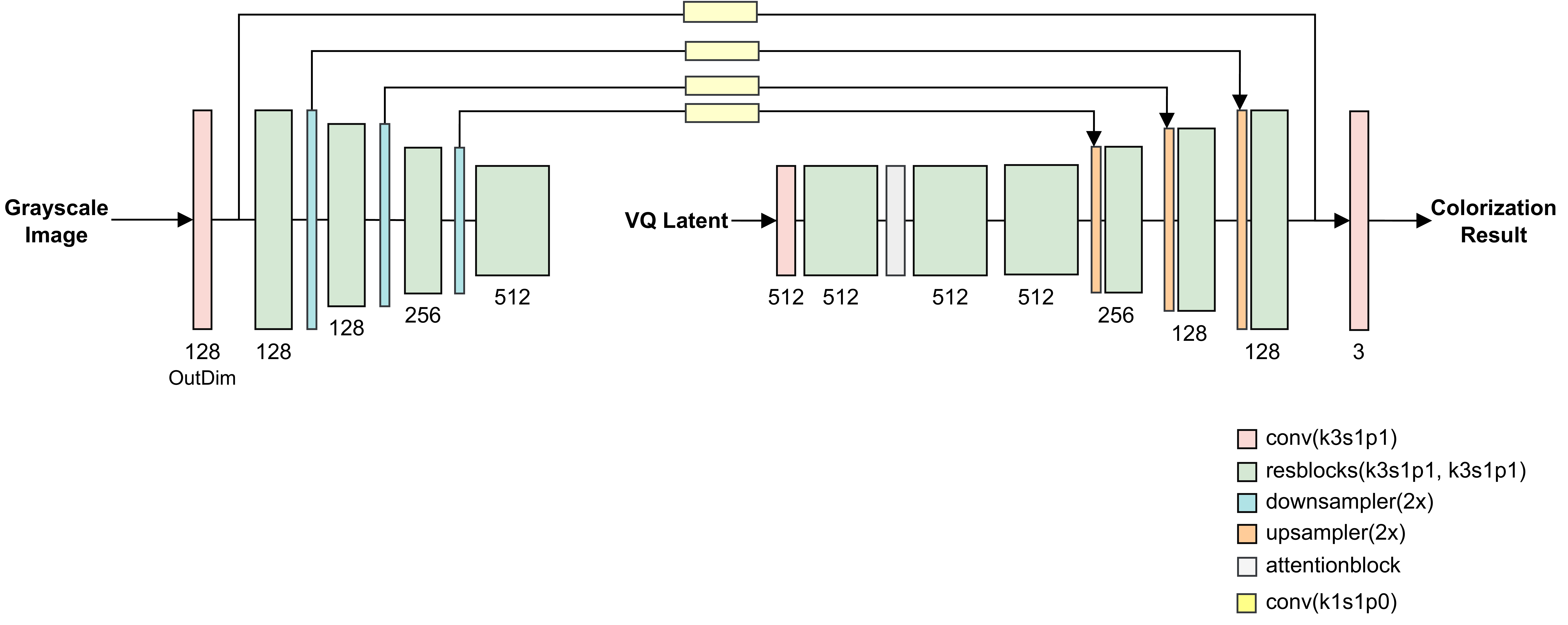}
\caption{Network architecture of \ColorizationDecoder.}
\label{fig:color-arch}
\end{figure}

The network architecture of the \ColorizationDecoder~is illustrated in Figure~\ref{fig:color-arch}. The grayscale encoder is a replica of the original VQVAE encoder but with the exclusion of the middle blocks. The \PriorModel~is a replica of the stable diffusion model as proposed by \cite{latent-diffusion}, but with modifications made to the input layers to facilitate the acceptance of concatenated inputs. The grayscale input and color hint map is encoded with a encoder using the same architecture of original VQVAE encoder. 

\end{document}